\documentclass{article}

     \PassOptionsToPackage{numbers, square}{natbib}

\usepackage[final]{neurips_2020}
\usepackage{amsfonts}       
\usepackage{amsthm}
\usepackage{amsmath}
\usepackage{adjustbox}
\usepackage{booktabs}       
\usepackage{capt-of}
\usepackage[T1]{fontenc}    
\usepackage{float}
\usepackage{graphicx}
\usepackage{hyperref}       
\usepackage[utf8]{inputenc} 
\usepackage{microtype}      
\usepackage{multirow}
\usepackage{nicefrac}       
\usepackage{subfigure}
\usepackage{url}            
\usepackage{xcolor}

\newcommand\eqdis{\mathrel{\overset{\makebox[0pt]{\mbox{\normalfont\tiny\sffamily d}}}{=}}}
\DeclareMathOperator*{\argmin}{arg\,min}
\DeclareMathOperator*{\argmax}{arg\,max}
\definecolor{rr}{RGB}{228,26,28}
\definecolor{bb}{RGB}{55,126,184}
\definecolor{gg}{RGB}{77,175,74}
\definecolor{pp}{RGB}{152,78,163}
\definecolor{oo}{RGB}{255,140,0}
\definecolor{new}{RGB}{204,102,0}

\title{Invertible Gaussian Reparameterization: Revisiting the Gumbel-Softmax}

\author{
  Andres Potapczynski \\
  Zuckerman Institute\\
  Columbia  University\\
  \texttt{ap3635@columbia.edu} \\
  \And
  Gabriel Loaiza-Ganem \\
  Layer 6 AI\thanks{Work partially done while at the Department of Statistics, Columbia University.}\\
  \texttt{gabriel@layer6.ai} \\
  \And
  John P. Cunningham \\
  Department of Statistics\\
  Columbia  University\\
  \texttt{jpc2181@columbia.edu}
}

\begin{document}

\maketitle

\begin{abstract}
The Gumbel-Softmax is a continuous distribution over the simplex that is often used as a relaxation
of discrete distributions. Because it can be readily interpreted and easily reparameterized, it
enjoys widespread use. We propose a modular and more flexible family of
reparameterizable distributions where Gaussian noise is transformed into a one-hot approximation
through an invertible function. This invertible function is composed of a modified softmax and can
incorporate diverse transformations that serve different specific purposes. For example, the
stick-breaking procedure allows us to extend the reparameterization trick to distributions with
countably infinite support, thus enabling the use of our distribution along nonparametric models, or normalizing flows let us increase the flexibility of the
distribution. Our construction enjoys theoretical advantages over the Gumbel-Softmax, such as closed
form $\mathbb{KL}$, and significantly outperforms it in a variety of experiments. Our code is available at \url{https://github.com/cunningham-lab/igr}.
\end{abstract}

\section{Introduction}
Numerous machine learning tasks involve optimization problems over discrete stochastic components
whose parameters we wish to learn. Mixture and mixed-membership models, variational autoencoders,
language models and reinforcement learning fall into this category \citep{johnson2016composing,
kingma2014auto, rezende2014stochastic, kusner2016gans, glynn1990likelihood}. Ideally, as with fully
continuous models, we would use stochastic optimization via backpropagation. One strategy to
compute the necessary gradients is using score estimators \citep{glynn1990likelihood,
williams1992simple}, however these estimates suffer from high variance which leads to slow
convergence. Another strategy is to find a
reparameterizable continuous relaxation of the discrete distribution. Reparameterization gradients
exhibit lower variance but are contingent on finding such a relaxation. \citet{jang2017categorical}
and \citet{maddison2017concrete} independently found such a continuous relaxation via the
Gumbel-Softmax (GS) or Concrete distribution.

The GS has experienced wide use and has been extended to other settings, such as permutations
\citep{linderman2018reparameterizing}, subsets \citep{xie2019reparameterizable} and more
\citep{balog2017lost}. Its success relies on several qualities that make it appealing: $(i)$ it is
reparameterizable, that is, it can be sampled by transforming parameter-independent noise through a
smooth function, $(ii)$ it can approximate any discrete distribution, (i.e. converge in
distribution) $(iii)$ it has a closed form density, and $(iv)$ its parameters can be interpreted
as the discrete distribution that it is relaxing. While the last quality is mathematically
pleasing, it is not a necessary condition for a valid relaxation. Here we ask: \emph{how important
is this parameter interpretability}? In the context of deep learning models, interpreting the
parameters is not a first concern, and we show that the GS can be significantly improved upon by
giving up this quality.

In this paper we propose an alternative family of distributions over the simplex to achieve this
relaxation, which we call Invertible Gaussian Reparameterization (IGR). Our reparameterization works
by transforming Gaussian noise through an invertible transformation onto the simplex, and a
temperature hyperparameter allows the distribution to concentrate its mass around the vertices. IGR
is more natural, more flexible, and more easily extended than the GS. Furthermore, IGR enables using the
reparameterization trick on distributions with countably infinite support, which enables nonparametric uses, and also admits closed
form $\mathbb{KL}$ divergence evaluation. Finally, we show that our distribution outperforms the GS
in a wide variety of experimental settings.

\section{Background}

\subsection{The reparameterization trick}

Many problems in machine learning can be formulated as optimizing parameters
of a distribution over an expectation:
\begin{equation}\label{gen_opt_prob}
\phi^* = \argmax_\phi L(\phi) := \argmax_\phi \mathbb{E}_{q_\phi(z)}[f(z)]
\end{equation}
where $q_\phi$ is a distribution over $\mathcal{S}$ parameterized by $\phi$ and $f: \mathcal{S}
\rightarrow \mathbb{R}$. In order to use stochastic gradient methods \citep{robbins1951stochastic,
bottou2012stochastic}, the gradient of $L$ has to be estimated. A first option is to use score
estimators \citep{glynn1990likelihood, williams1992simple}. However, in practice score estimators
usually exhibit high variance \citep{miller2017reducing}. The reparameterization trick
\citep{kingma2014auto} provides an alternative estimate of this gradient which empirically has less
variance, resulting in more efficient optimization. The reparameterization trick consists of finding
a function $g(\epsilon, \phi)$ such that $g$ is differentiable with respect to $\phi$ and if $z\sim
q_\phi$, then:
\begin{equation}\label{rep_trick}
 z \eqdis g(\epsilon, \phi)
\end{equation}
where $\epsilon$ is a continuous random variable whose distribution does not depend on $\phi$ and is
easy to sample from. The gradient is then estimated by:
\begin{equation}\label{rep_grad}
 \nabla_\phi L(\phi) \approx \dfrac{1}{B} \displaystyle \sum_{b=1}^B \nabla_\phi f(g(\epsilon_b,
 \phi))
\end{equation}
where $\epsilon_1,\dots, \epsilon_B$ are iid samples from the distribution of $\epsilon$. For
example, if $\phi=(\mu, \sigma)$ and $q_\phi = \mathcal{N}(\mu, \sigma^2)$ then the
reparameterization trick is given by $g(\epsilon, \phi)=\mu + \sigma\epsilon$ with $\epsilon \sim
\mathcal{N}(0,1)$.

\subsection{Continuous relaxations}\label{sec:relax}

While we can use score estimators whether $q_\phi$ has
continuous or discrete support, the reparameterization gradient of equation \ref{rep_grad} is only
valid when $q_\phi$ has continuous support. To extend the reparameterization trick to the discrete
setting, thus avoiding the high variance issues of score estimators, suppose $q_\phi$ is a
distribution over the set $\mathcal{S}=\{1,2,\dots, K\}$. We use one-hot representations of length
$K$ for the elements of $\mathcal{S}$, so that $\mathcal{S}$ can be interpreted as the vertices of
the $(K-1)$-simplex, $\Delta^{(K-1)}=\{z \in \mathbb{R}^K: z_k \geq 0 \text{ and }\sum_{k=1}^K
z_k=1\}$. The idea is to now place a continuous distribution over $\Delta^{(K-1)}$ that approximates $q_\phi$.
Note that placing a distribution over $\Delta^{(K-1)}$ is equivalent to placing a distribution over
$\mathcal{S}^{(K-1)}=\{z \in \mathbb{R}^{K-1}: z_k > 0 \text{ and }\sum_{k=1}^{K-1} z_k <
1\}$, as the last coordinate can
be obtained from the previous ones: $z_K = 1-\sum_{k=1}^{K-1}z_k$. Placing a distribution over
$\mathcal{S}^{(K-1)}$ is mathematically convenient as $\mathcal{S}^{(K-1)} \subset \mathbb{R}^{K-1}$
has positive Lebesgue measure, while $\Delta^{(K-1)} \subset \mathbb{R}^K$ does not. Although this
distinction might appear as an irrelevant technicality, it allows us to correctly compute our Jacobians
in section \ref{gaudi}. We will thus interchangeably think of distributions over $\mathcal{S}$ as points in either $\mathcal{S}^{(K-1)}$ or $\Delta^{(K-1)}$. The optimization problem of equation
\ref{gen_opt_prob} is then relaxed to:

\begin{equation}\label{relax_opt_prob}
 \tilde{\phi}^* = \argmax_{\tilde{\phi}} \tilde{L}(\tilde{\phi}) :=
 \argmax_{\tilde{\phi}}\mathbb{E}_{\tilde{q}_{\tilde{\phi}}(\tilde{z})}[\tilde{f}(\tilde{z})]
\end{equation}

where $\tilde{q}_{\tilde{\phi}}$ is a distribution over $\mathcal{S}^{(K-1)}$ and the function
$\tilde{f}:\mathcal{S}^{(K-1)} \rightarrow \mathbb{R}$ is a relaxation of $f$ to
$\mathcal{S}^{(K-1)}$. As long as $\tilde{q}_{\tilde{\phi}}$ concentrates most of its mass around
$\mathcal{S}$ and $\tilde{f}$ is smooth, this relaxation is sensible. If $\tilde{q}_{\tilde{\phi}}$
can be reparameterized as in equation \ref{rep_trick}, then we can use the reparameterization trick.
We make two important notes: first, not only the distribution is relaxed, the function $f$ also has
to be relaxed to $\tilde{f}$ because it now needs to take inputs in $\mathcal{S}^{(K-1)}$ and not
just $\mathcal{S}$. In other words, the objective must also be relaxed, not just the distribution.
Second, the parameters $\tilde{\phi}$ of the relaxed distribution need not match the parameters
$\phi$ of the original distribution.

\citet{maddison2017concrete} and \citet{jang2017categorical} proposed the Gumbel-Softmax
distribution, which is parameterized by $\alpha \in (0,\infty)^K$ and a temperature hyperparameter
$\tau > 0$, and is reparameterized as:
\begin{equation}\label{GS_rep}
 \tilde{z} \eqdis \text{softmax}\big((\epsilon + \log \alpha) / \tau \big)
\end{equation}
where $\epsilon \in \mathbb{R}^K$ is a vector with independent $\text{Gumbel}(0,1)$ entries and
$\log$ refers to elementwise logarithm. Note that when the temperature approaches $0$, not only
does the GS concentrate its mass around $\mathcal{S}$, but it converges to a distribution
proportional to $\alpha$. The GS distribution implied by equation \ref{GS_rep} can be shown to be:
\begin{equation}\label{GS_distr}
 \tilde{q}_{\alpha, \tau}(\tilde{z}) = (K-1)! \,\, \tau^{K-1}\displaystyle \prod_{k=1}^K
 \left(\dfrac{\alpha_k \tilde{z}_k^{-\tau - 1}}{\sum_{j=1}^K \alpha_j \tilde{z}_j^{-\tau}}\right)
\end{equation}
We highlight the difference between $\alpha$ and $\tilde{\phi}$: the former is the parameter of the
GS distribution and might depend on the latter, which is the parameter of the loss with respect to which we
optimize in equation \ref{relax_opt_prob}. For example, $\alpha$ might be the output of a neural
network parameterized by $\tilde{\phi}$. A common use of
the GS is to relax objectives of the form:
\begin{equation}
 \mathbb{KL}(q_\phi||p_0) = \mathbb{E}_{q_\phi(z)}\Big[\log \dfrac{q_\phi(z)}{p_0(z)}\Big]
\end{equation}
where $p_0$ is a distribution over $\mathcal{S}$. Relaxing this $\mathbb{KL}$ requires additional
care: it cannot be relaxed to $\mathbb{KL}(\tilde{q}_{\tilde{\phi}}||p_0)$ because the $\mathbb{KL}$
divergence is not well defined between a continuous and a discrete distribution. In other words,
relaxing $f$ to $\tilde{f}$ is not straightforward when a $\mathbb{KL}$ divergence is involved in
the objective. When using a GS relaxation, researchers commonly replace this $\mathbb{KL}$
with \citep{jang2017categorical, dupont2018learning, tucker2017rebar}:
\begin{equation}\label{relax_kl_bad}
 \mathbb{KL}(\bar{\alpha}||p_0) \text{ where }\bar{\alpha}_k=\dfrac{\alpha_k}{\sum_{i=1}^K
 \alpha_i}
\end{equation}
the idea being that, for low temperatures, the GS approximates a distribution proportional to its
parameter, i.e. $\bar{\alpha} \in \Delta^{(K-1)}$. The goal of this substitution is to still compute a $\mathbb{KL}$
between two discrete variables, even after relaxing the distribution. This substitution is
problematic, as pointed out by \citet{maddison2017concrete}, as it does not take into account how
close the GS actually is to $\bar{\alpha}$. A more sensible way to relax the discrete $\mathbb{KL}$
is to relax it to an actual continuous $\mathbb{KL}$ as done by \citet{maddison2017concrete}:
\begin{equation}\label{relax_kl}
 \mathbb{KL}(\tilde{q}_{\tilde{\phi}}|| \tilde{q}_0)
\end{equation}
where $\tilde{q}_0$ is fixed in such a way that it is close to $p_0$.
For the GS, finding such a distribution is straightforward as a consequence of its parameter
interpretability: $\tilde{q}_0$ can be chosen as a GS with parameter $\alpha_0 =
p_0$. Note that the $\mathbb{KL}$ in equation \ref{relax_kl} cannot be directly evaluated,
but a Monte Carlo estimate can be formed thanks to the closed form density of equation
\ref{GS_distr} and thus stochastic gradient descent can be performed.

Finally, it is worth remarking that while \citet{stirn2019new} and
\citet{gordon2020continuous} proposed distributions over the simplex which admit
reparameterization gradients, their goals are not to obtain discrete relaxations. Thus
they do not have a temperature hyperparameter allowing to concentrate mass on the
vertices to approximate discrete distributions.

\section{The invertible Gaussian reparameterization family} \label{gaudi}
If the only requirements for a continuous relaxation are a reparameterizable distribution
on the simplex and a temperature hyperparameter allowing to concentrate mass around the
vertices, one might logically ask: why use the specific choices of the GS? Namely, why
use the unusual Gumbel noise and be forced to use the softmax as a mapping onto the
simplex? If tasked with constructing a reparametrizable distribution on the simplex, we
argue that the most natural choice is to sample Gaussian noise and map it to the simplex;
trying different mappings and keeping the best performing one. The cost of this choice is
losing the parameter interpretability of the GS, but we will show the advantages are
numerous and well worth this cost.

 We now present the IGR distribution on $\mathcal{S}^{(K-1)}$, which is parameterized by $(\mu, \sigma)$, where $\mu \in \mathbb{R}^{K-1}$ and
 $\sigma \in (0, \infty)^{K-1}$. Gaussian noise $\epsilon = (\epsilon_1,\dots, \epsilon_{K-1}) \sim
 \mathcal{N}(0, I_{K-1})$ is transformed in the following way:

 \begin{align}
 y & = \mu + \text{diag}(\sigma) \epsilon \\ \tilde{z} & = g(y, \tau)
 \end{align}

 where $\text{diag}(\sigma)$ is a diagonal matrix whose nonzero elements are given by $\sigma$,
 $g(\cdot, \tau)$ is an invertible smooth function and $\tau > 0$ is a temperature hyperparameter.
 Note that IGR is not only more natural than the GS, but is is also more flexible, having
 $2K-2$ parameters instead of $K$. The first advantage of choosing $g$ to be an invertible function
 is that the density of $\tilde{z}$ can be computed in closed form with the change of variable
 formula:

 \begin{equation}
 \tilde{q}_{\mu, \sigma, \tau}(\tilde{z}) = \mathcal{N}(y|\mu, \sigma)|\det J_{g}(y,
 \tau)|^{-1}
 \end{equation}

 where $J_{g}(\cdot, \tau)$ is the Jacobian of $g(\cdot, \tau)$. The second advantage of this choice
 is that it allows us to compute the $\mathbb{KL}$ in closed form (as the Jacobian terms cancel out
 in the ratio):
\begin{equation}
  \mathbb{KL}\left(\tilde{q}_{\mu, \sigma, \tau}(\tilde{z})||\tilde{q}_{\mu_0, \sigma_0, \tau}(\tilde{z})\right)
  =
  \mathbb{KL}\left(\mathcal{N}(\mu, \sigma^2)||\mathcal{N}(\mu_0, \sigma^2_0)\right)
\end{equation}
 and thus Monte Carlo estimation of equation \ref{relax_kl} is no longer needed.

 The components of the IGR can be easily mixed-and-matched. For example, while we use Gaussian noise as the most natural first choice because it is reparameterizable and because the
$\mathbb{KL}$ divergence between two Gaussians has closed form, any other choice with these two
properties can also be used. Similarly, any choice of $g$, as long as it obeys some requirements which we explain in the section \ref{g}, can also be used. In contrast, changing the Gumbel distribution or the softmax used in the GS cannot be done. These properties of the IGR make it more easily extensible than the GS.

Since the parameter
 interpretability of the GS is lost in IGR, we cannot directly read $\mu_0$ and $\sigma_0$ from
 $p_0 \in \mathcal{S}^{(K-1)}$. Thus when a $\mathbb{KL}$ term is involved, while IGR gains the ability to evaluate it
 analytically, we solve the following optimization problem to obtain these parameters:

 \begin{equation} \label{opt_prior}
 (\mu_0, \sigma_0) = \argmin_{(\mu, \sigma)} \mathbb{E}_{\tilde{q}_{\mu, \sigma,
 \tau}(\tilde{z})}[||\tilde{z} - p_0||^2_2]
 \end{equation}

 Note that having to solve this problem is a very small price to pay for losing parameter
 interpretability: the optimization is a very simple moment matching problem and has to be be
 computed only once for any given $p_0$.

 \subsection{Choosing $g(\cdot, \tau)$} \label{g}
 In this section we design some invertible functions that could be used and argue the rationale
 behind their construction. There are two important desiderata for $g(\cdot, \tau)$: the first one
 is that we should be able to compute the determinant of its Jacobian efficiently, which enables
 tractable density evaluation. This tractability can be achieved, for example, by ensuring the
 Jacobian is triangular. Note that although in many instances we do not actually require evaluating
 the density of the relaxation (e.g. variational autoencoders \citep{kingma2014auto}), this is a
 problem-specific property and density evaluation remains desirable in general. The second is that
 the limit as $\tau \rightarrow 0$ of $g(y, \tau)$ is in $\mathcal{S}$ for almost all $y$, meaning
 that as the temperature gets smaller, the distribution places most of its mass around the vertices.
 The two most natural choices for mapping onto the simplex are the softmax function and the
 stick-breaking procedure.  As we explain below, these alone are not enough, and we thus modify them
 to make them appropriate for our purposes. The softmax has two issues: first, it maps to
 $\Delta^{(K-1)}$ and not $\mathcal{S}^{(K-1)}$ and second, it is not invertible. Both of these
 problems can be addressed with a small modification of the softmax function:

 \begin{equation}\label{softmaxpp}
  \begin{split}
    g(y, \tau)_k = \frac{\exp(y_k / \tau)} {\sum_{j=1}^{K-1} \exp(y_j / \tau)) + \delta}
  \end{split}
 \end{equation}
 where $\delta > 0$ ensures that the function is invertible and maps to $\mathcal{S}^{(K-1)}$.
 Furthermore, the Jacobian of this transformation can be efficiently computed with the matrix
 determinant lemma (see appendix for details). We will refer to this transformation as the
 softmax$_{++}$.

 The other natural alternative to map from $(0, 1)^{K-1}$ onto $\mathcal{S}^{(K-1)}$ is through the
 stick-breaking procedure \citep{ferguson1973bayesian}, which we briefly review here. Given $u \in
 (0, 1)^{K-1}$, the result $v = SB(u)$ of performing stick-breaking on $u$ is given by:
 \begin{equation}
 v_k = \displaystyle u_k \prod_{i=1}^{k-1} (1-u_i), \text{ for }k=1,2,\dots,K-1
 \end{equation}

 In addition to producing outputs in $\mathcal{S}^{(K-1)}$, this procedure has some useful
 properties, namely: it is invertible, its Jacobian is triangular, and it can easily be extended to
 the case where $K=\infty$ (which will be useful to extend IGR to relax discrete distributions with
 countably infinite support). While the invertibility property might suggest that the
 stick-breaking procedure alone is enough to use with IGR, a temperature hyperparameter $\tau$
 still needs to be introduced in such a way that as $\tau \rightarrow 0$, the resulting
 distribution concentrates its mass on the vertices. Unlike with the $\text{softmax}_{++}$, simply
 dividing the input by $\tau$ does not achieve this limiting behavior. The most natural way of introducing a
 temperature that achieves the desired limiting behavior is by linearly interpolating to the nearest
 vertex, resulting in a $g$ function given by:

 \begin{equation}\label{sb_proj}
 \begin{cases}
  w  = SB\big(\text{sigmoid}(y)\big)\\
  g(y, \tau)  = \tau w + (1-\tau)P_\mathcal{S}(w)
 \end{cases}
 \end{equation}
 where $P_{\mathcal{S}}$ is the projection onto the vertices of $\mathcal{S}^{(K-1)}$. Note that
 the Jacobian of this transformation is triangular. However, we found better empirical performance
 with the following function, which introduces the temperature using the $\text{softmax}_{++}$
 function:
 \begin{equation}\label{infinite_g}
 g(y, \tau) = \text{softmax}_{++}(w , \tau)
 \end{equation}
 While it might seem redundant to apply both a stick-breaking procedure and a $\text{softmax}_{++}$
 as they both map to $\mathcal{S}^{(K-1)}$, the $\text{softmax}_{++}$ function allows to introduce
 $\tau$ in such a way that the distribution concentrates its mass around the vertices as $\tau
 \rightarrow 0$. Also, as seen in section \ref{inf}, the stick-breaking procedure proves
 useful as it enables using the
 reparameterization trick in the countably infinite support setting.

 Finally, another choice of $g(\cdot, \tau)$ could be a normalizing flow
 \citep{rezende2015variational, kingma2016improved, dinh2017density} followed by
 $\text{softmax}_{++}$. Normalizing flows are flexible neural networks constructed in
 such a way that they are invertible while still allowing tractable Jacobian determinant
 evaluations, so that they enable us to learn $g$. We note that normalizing flows require additional
 parameters, so that when using them, IGR is not only parameterized by $\mu$ and $\sigma$, but by
 the parameters of the normalizing flow as well. Thus, if a $\mathbb{KL}$ is involved, the
 optimization problem of equation \ref{opt_prior} needs to be solved over the parameters of the
 normalizing flow too, and as a result the $\mathbb{KL}$ in equation \ref{relax_kl} cannot be
 evaluated in closed form anymore, as the parameters of the two involved normalizing flows need not
 match. However, Monte Carlo estimates of the $\mathbb{KL}$ are still readily available.

 \subsection{Reparameterization trick for countably infinite distributions} \label{inf}

 Since the stick-breaking procedure can map to $\mathcal{S}^{\infty} = \{z \in \mathbb{R}^\infty
 : z_k \geq 0 \text{ and }\sum_{k=1}^\infty z_k=1\}$, we can extend equation
 \ref{infinite_g} to the setting where the discrete distribution has countably infinite
 support (e.g. Poisson, geometric or negative binomial distributions). In this setting, the IGR
 is parameterized by $\mu \in \mathbb{R}^\infty$ and $\sigma \in (0, \infty)^\infty$. Clearly
 backpropagating through infinitely many parameters cannot be done in a computer, but we do not
 have to do so as most of the parameters contribute very little to the loss. For a sample
 $\epsilon_1, \epsilon_2, \dots$ we only update the first $K$ coordinates of $\mu$ and $\sigma$,
 where $K$ is the number such that:

 \begin{equation}\label{truncation}
  \begin{split}
    \sum_{k=1}^{K-1} g(y, \tau)_k \leq \rho < \sum_{k=1}^{K} g(y, \tau)_k
  \end{split}
 \end{equation}

 where $\rho \in (0, 1)$ is a pre-specified precision hyperparameter and $g$ is as in equation
 \ref{infinite_g}. Note that here $K$ is now a random variable that depends on $\epsilon$ instead of
 being fixed as before, so that in a way the number of (effective) categories gets learned by the
 data. Note as well that the stick-breaking procedure is necessary to know where to cut $K$ as it
 guarantees that later terms in the sequence are small, which would not happen if we only applied a
 $\text{softmax}_{++}$ function.

 \subsection{Recovering the discrete distribution}\label{recovering}

 Recall that the original objective of continuous relaxations is to solve the discrete problem of
 equation \ref{gen_opt_prob}, so that once we have solved the continuous problem of equation
 \ref{relax_opt_prob}, it is desirable to have the ability to recover a solution to the former
 problem. In other words, given the parameters of a continuous relaxation, we should be able to
 recover the discrete distribution that it is relaxing. The parameter intepretability of the GS
 allows to directly do so. In this section we derive a method for doing so with the IGR, which is
 enabled by the following proposition:\\

 \textbf{Proposition 1}: For any $\delta > 0$, the following holds:
 \begin{equation}
 \lim_{\tau \rightarrow 0} \text{softmax}_{++}(y , \tau) = h(y) := \begin{cases}
 e_{k^*}, \text{ if }k^*=\displaystyle \argmax_{k=1,\dots,K-1}(y_k)\text{ and }\max_{k=1,\dots,K-1}(y_k) > 0\\
 0\hspace{9pt},\text{ if }\displaystyle \max_{k=1,\dots,K-1}(y_k) < 0
 \end{cases}
 \end{equation}
 where $e_k \in \mathbb{R}^{K-1}$ is the one-hot vector with a $1$ in its $k$-th coordinate.\\
 \textbf{Proof}: See appendix.\\

Thus, the vector of discrete probabilities associated with IGR is
$\mathbb{E}[h(\tilde{z})]$, which can be easily approximated through a Monte Carlo estimate by
sampling from the IGR and averaging the results after transforming them with $h$. This is the last
cost to pay for losing parameter interpretability, but once again it is very small: the complexity
of this approximation is negligible when compared to the one of solving the problem of equation
\ref{relax_opt_prob}. Note also that this proposition enables the use of straight-through estimators \citep{bengio2013estimating}, where the sample is discretized during the forward pass, but not for backpropagation. The next proposition shows that when just using the $\text{softmax}_{++}$ as
$g$, the recovered discrete distribution can be written in an even more explicit form:\\

\textbf{Proposition 2}: If $y_k \sim \mathcal{N}(\mu_k, \sigma_k)$ for $k=1,\dots, K-1$, and we
define the discrete random variable $H$ by $H = k$ if $h(y)=e_k$ and $H = K$ if $h(y)=0$, then:
\begin{equation}\label{discrete_recovered}
\mathbb{P}(H=k)=\begin{cases}
\displaystyle \int_{0}^\infty \dfrac{1}{\sigma_k}\phi\left(\dfrac{t - \mu_k}{\sigma_k}\right) \prod_{j \neq k}
\Phi\left(\dfrac{t - \mu_j}{\sigma_j}\right) dt ,\text{ if }k=1,\dots,K-1\\
\displaystyle \prod_{j=1}^{K-1} \Phi\left(-\dfrac{\mu_j}{\sigma_j}\right) \hspace{101pt}, \text{ if }k=K
\end{cases}
\end{equation}
where $\phi$ and $\Phi$ are the standard Gaussian density and cumulative distribution function, respectively.\\
 \textbf{Proof}: See
appendix.\\

We finish this section by noting that there is literature proposing gradient
estimators and attempting to reduce their variance \citep{mnih2016variational,
miller2017reducing, tucker2017rebar, grathwohl2017backpropagation, kool2020estimating}.
In particular, \citet{grathwohl2017backpropagation} and \citet{tucker2017rebar} proposed
techniques involving the GS. Their techniques, however, require computing the gradient of
the discrete objective with respect to the parameters of the continuous relaxation, which
can be done with the GS thanks to its parameter interpretability.  Proposition 2 thus
enables the use of their methods with IGR, as the integral in equation
\ref{discrete_recovered} can be easily approximated numerically. Due to space constraints
we include details, along a discussion about bias, in the appendix.

\section{Experiments} \label{exp}

In this section, we contrast the performance of the IGR (with different choices of $g$) alongside
that of the GS. First, in relation to section \ref{inf}, we compare the ability of the IGR and the
GS (with varying number of categories) to approximate a countably infinite distribution.  We then
focus on tasks that involve a $\mathbb{KL}$ term in their objective function. Finally, we also
consider a Structured Output Prediction task which does not involve a $\mathbb{KL}$ term.  For the
experiments involving a $\mathbb{KL}$ term, we use variational autoencoders (VAEs)
\cite{kingma2014auto}. We follow the setup of \citet{maddison2017concrete} and \citet{jang2017categorical}
(although note that we use a slightly different objective than \citet{jang2017categorical}, see
appendix for details). The datasets we use are handwritten digits from MNIST, fashion items from
FMNIST and alphabet symbols from Omniglot. We ran each experiment 5 times and report averages
plus/minus one standard deviation.  Additionally, for all the experiments, we used the log scale
implementation of the GS (ExpConcrete) as in \citet{maddison2017concrete} since it avoids numerical
issues and allows us to run the models involving the GS at lower temperatures. Throughout this
section, the label IGR-I denotes the implementation with the $\text{softmax}_{++}$ (equation
\ref{softmaxpp}), the label IGR-SB the implementation with the stick-breaking transformation
followed by a $\text{softmax}_{++}$ (equation \ref{infinite_g}), and finally the label IGR-Planar the
implementation using two nested Planar Flows \citep{rezende2015variational} followed by a
$\text{softmax}_{++}$ .

Comparing any IGR variant against the GS requires selecting temperature hyperparameters for each
model. To make a fair comparison, temperatures $\tau$ should be chosen carefully as they affect
models differently, so they cannot just be set to the same value. We thus choose the temperature
hyperparameter through cross validation, considering the range of possible temperatures $\{0.01,
0.03, 0.07, 0.1, 0.25, 0.4, 0.5, 0.67, 0.85, 1.0\}$ and compare best-performing models. However, and
very importantly, we use the loss on the recovered \textit{discrete} model --- not the trained
\textit{continuous} one --- to select the best performing model. This avoids the potential issue of
having one model produce better discrete relaxations which are closer to the vertices of the
simplex, while resulting in a larger continuous loss as the other model is allowed to use the
simplex more freely.  All implementation details are in the appendix.

 \subsection{Approximating a Poisson Distribution}

\begin{figure}
\centering
\begin{tabular}{cc}
\includegraphics[scale=0.4]{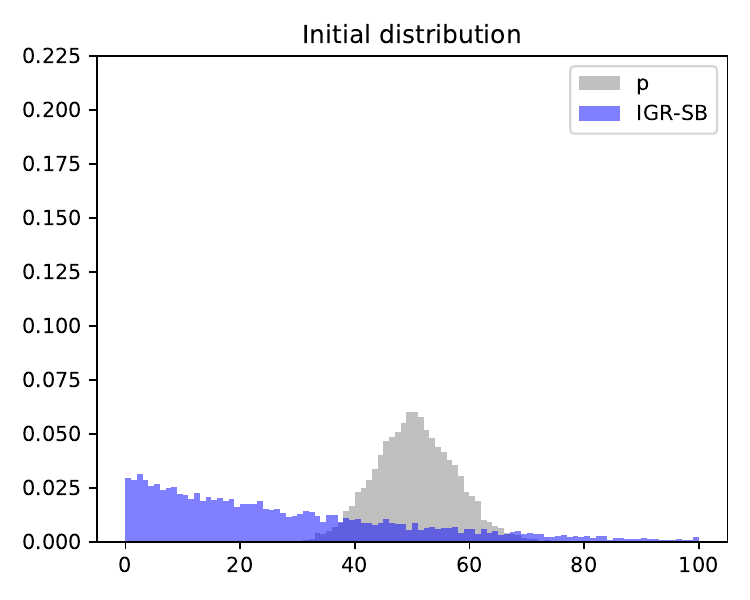}&
\includegraphics[scale=0.4]{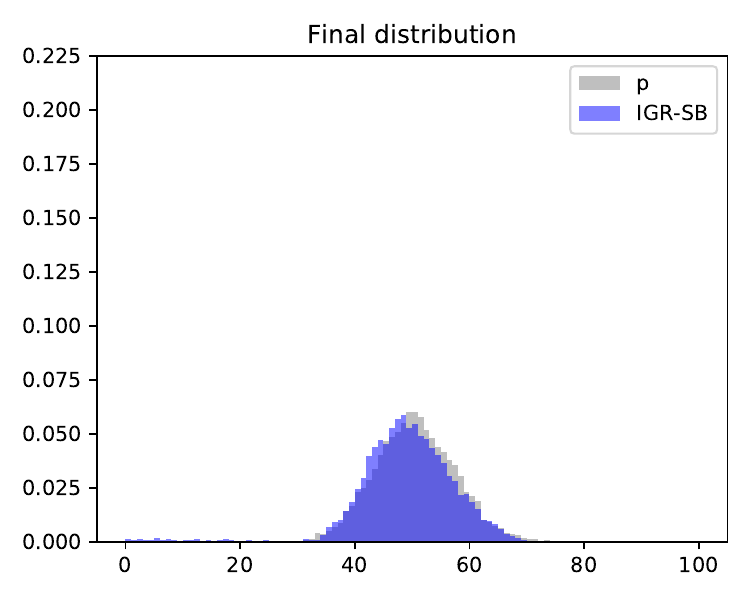}\\
\includegraphics[scale=0.4]{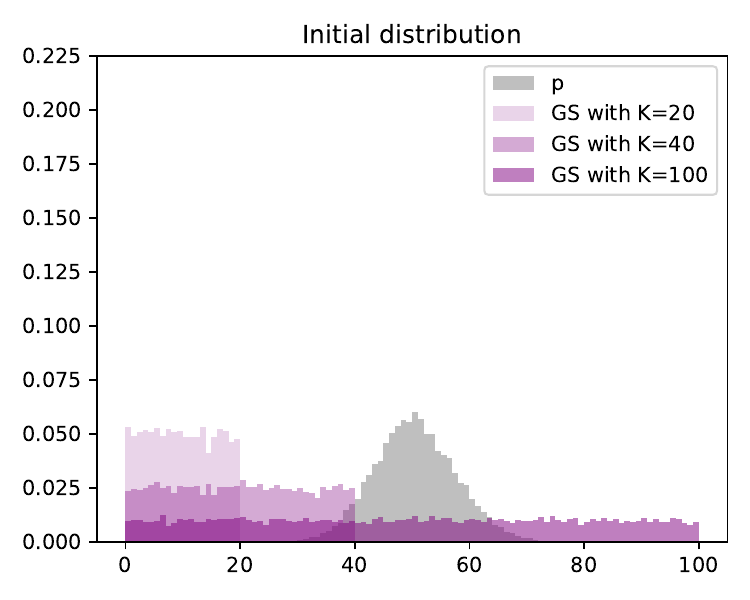}&
\includegraphics[scale=0.4]{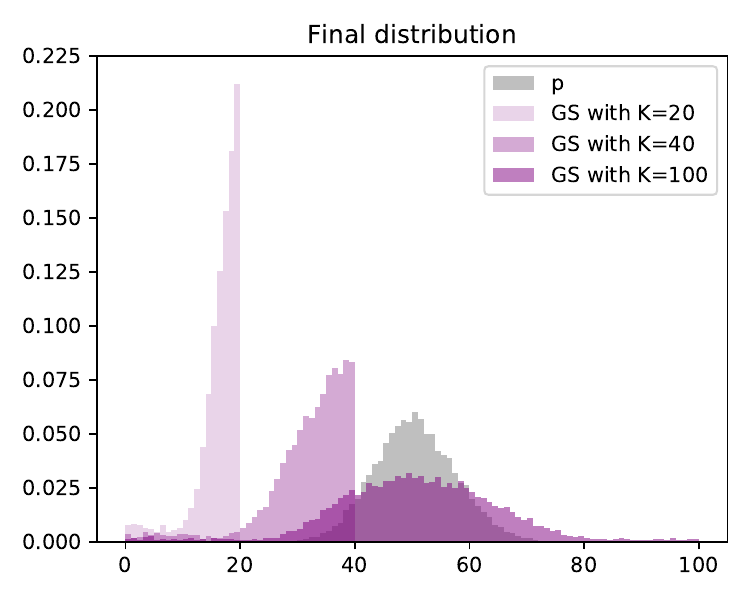}
\end{tabular}
\caption{Approximations to a Poisson distribution for IGR-SB (top right panel) and GS (bottom right
panel) after 1,000 training steps. Initial values of the approximations are displayed on the
respective left panels. Due to the stick-breaking procedure, a random initialization
concentrates to the left.}\label{fig:poisson}
\end{figure}

 Here we compare the ability of the IGR-SB and the GS to approximate distributions with countably
 infinite support. The top panels of Figure \ref{fig:poisson} show an approximation with the IGR-SB
 to a Poisson distribution with $\lambda=50$, while the bottom panels show the same approximations
 when using a GS with different number $K$ of discrete components. These approximations are computed
 by optimizing the objective in equation \ref{opt_prior}. We can see how the IGR-SB outperforms the
 GS without having to specify $K$. We show further comparisons when approximating other
 distributions in the appendix.

 \subsection{Variational Autoencoders}

 We trained VAEs composed of $20$ discrete variables with $10$ categories each. VAEs are latent
 variable models which maximize the ELBO, a lower bound on the log likelihood involving a
 $\mathbb{KL}$ term (see appendix for details). For MNIST and Omniglot we used a fixed binarization
 and a Bernoulli decoder, whereas for FMNIST we use a Gaussian decoder. Table \ref{rest_cat} shows
 test log-likelihoods (not ELBOs, these are obtained as in \citet{burda2015importance} with
 $m=1000$, and are computed on the recovered discrete model) plus/minus one standard deviation for two different architectures. We highlight best results and those within error. The
 IGR performs best or is within error, except in a single scenario. We report the test log-likelihood as it is the most relevant metric from a machine learning perspective, but from an optimization point of view, the discretized training ELBO is of more interest, as it more accurately measures how well the original objective is being maximized. We include this evaluation, which is also favorable to IGR, in the appendix. It is also worth mentioning that the execution times between the IGR and the GS were
 almost identical for the I and SB variants. Nonetheless, the IGR Planar is about $30\%$ slower than
 all the other alternatives.

 To verify how much of our performance improvement is due to our closed form $\mathbb{KL}$, we also
 trained the VAE using the \emph{sticking the landing} gradient estimator proposed by
 \citet{roeder2017sticking}, which does not involve a closed form $\mathbb{KL}$ divergence. Results are also
 shown in Table \ref{rest_cat} (with the label SL). Note that all SL models outperform their non-SL
 counterparts, suggesting that the closed form $\mathbb{KL}$ of the IGR is not a key component of
 its superior empirical performance. We note that closed form $\mathbb{KL}$ remains an attractive
 theoretical property which could prove more useful in other applications.

 In Figure \ref{mnist_cat} we show that the IGR also outperforms the GS on the continuous model (not
 only the discrete one). The plot contains error bars, but these are almost imperceptible due to
 their size and the scale of the plot. Note that while we include this comparison for completeness,
 as we believe that the most relevant comparisons are on the recovered discrete model, it is
 interesting to see that the performance gains of the IGR over the GS on discretized models do not
 come at the cost of poorer continuous ones.

 \begin{table}[]
 \begin{center}
 \scalebox{0.9}{
 \begin{tabular}{p{2.5cm}lp{2.5cm}p{2.5cm}p{2.5cm}}
 \textbf{Architecture}&\textbf{Discrete Models} & \textbf{MNIST} & \textbf{FMNIST} & \textbf{Omniglot} \\
 \hline
 \hline
\multirow{4}{*}{Linear} & \textcolor{gg}{\textbf{IGR-I}}            & \textbf{-94.65 $\pm$  0.14}           & \textbf{-38.12 $\pm$ 0.12}           & -128.14 $\pm$ 0.40 \\
 & \textcolor{rr}{\textbf{IGR-Planar}}       & -96.21 $\pm$  0.14           & -38.72 $\pm$ 0.17           & -130.76 $\pm$ 0.17 \\
 & \textcolor{bb}{\textbf{IGR-SB}}           & -96.74  $\pm$ 0.36           & -41.70 $\pm$ 0.50           & \textbf{-124.77 $\pm$ 0.40}\\
 \vspace{0.05cm}
 & \textcolor{pp}{\textbf{GS}}               & -106.17 $\pm$ 1.00           & -46.65 $\pm$ 0.89           & -138.98 $\pm$ 1.01 \\
 \hline
\multirow{4}{*}{Non-linear} & \textcolor{gg}{\textbf{IGR-I}} & \textbf{-91.98 $\pm$ 1.29} & \textbf{-34.80 $\pm$ 3.33} & \textbf{-135.30 $\pm$ 1.71}\\
 & \textcolor{rr}{\textbf{IGR-Planar}} & \textbf{-92.91 $\pm$ 2.51} & \textbf{-34.10 $\pm$ 3.23} & \textbf{-133.63 $\pm$ 1.86}\\
 & \textcolor{bb}{\textbf{IGR-SB}} & -94.92 $\pm$ 0.66 & \textbf{-34.57 $\pm$ 3.09} & \textbf{-139.82 $\pm$ 9.27}\\
 & \textcolor{pp}{\textbf{GS}} & -98.06 $\pm$ 1.73 & \textbf{-29.72 $\pm$ 2.77} & -147.71 $\pm$ 3.04\\
  \hline
\multirow{4}{*}{Linear} & \textcolor{gg}{\textbf{IGR-I + SL}}       & \textbf{-94.18 $\pm$ 0.37}   & \textbf{-38.16 $\pm$ 0.35}  & \textbf{-122.96 $\pm$ 1.32} \\
 & \textcolor{rr}{\textbf{IGR-Planar + SL}}  & -95.97 $\pm$ 0.53            & \textbf{-38.59 $\pm$ 0.29}           & \textbf{-127.96 $\pm$ 3.75} \\
 & \textcolor{bb}{\textbf{IGR-SB + SL}}      & -96.05 $\pm$ 0.74            & -39.52 $\pm$ 0.32           & \textbf{-124.35 $\pm$ 1.10} \\
 & \textcolor{pp}{\textbf{GS + SL}}          & -103.80 $\pm$ 0.73           & -43.86 $\pm$ 1.22           & -133.45 $\pm$ 1.88 \\
 \hline
 \multirow{4}{*}{Non-Linear}
 & \textcolor{gg}{\textbf{IGR-I + SL}}
 & -91.38 $\pm$ 0.86            & -34.39 $\pm$ 0.67           & -134.60 $\pm$ 0.68\\
 & \textcolor{rr}{\textbf{IGR-Planar + SL}}
 & \textbf{-88.81 $\pm$ 0.49}   & -33.99 $\pm$ 1.82           & \textbf{-129.47 $\pm$ 1.06}\\
 & \textcolor{bb}{\textbf{IGR-SB + SL}}
 & -92.67 $\pm$ 1.48            & -34.86 $\pm$ 1.30           & -135.82 $\pm$ 2.58\\
 & \textcolor{pp}{\textbf{GS + SL}}
 & -97.87 $\pm$ 0.61            & \textbf{-28.81 $\pm$ 0.64}  & -140.37 $\pm$ 0.25\\
 [\smallskipamount]
 \end{tabular}}
 \caption{Test log-likelihood on MNIST, FMNIST and Omniglot for IGR and GS. Higher is better.}
 \label{rest_cat}
 \end{center}
 \end{table}

 Finally, we compared IGR and the GS using the variance reduction technique of
 \citet{grathwohl2017backpropagation}, whose use is enabled thanks to proposition 2.
 We include this comparison --- which was yet again favorable to IGR --- and the corresponding
 discussion, along with a comparison against the estimator proposed by \citet{kool2020estimating}, in the appendix.

  \subsection{Structured Output Prediction}
  We consider a Structured Output Prediction task, where we reconstruct the lower part of
  an image given the upper part by using a binary stochastic feedforward neural network.
  In contrast to our previous experiments, this is a task that does not require the
  computation of a $\mathbb{KL}$ divergence. It was first proposed in
  \citet{raiko2014techniques} and replicated by \citet{gu2015muprop},
  \citet{jang2017categorical} and \citet{maddison2017concrete}, and does not involve a
  $\mathbb{KL}$. The results of this experiment are in Table \ref{ap:sop}, where we can
  see that once again, IGR outperforms the GS.

\begin{table}[t]
\begin{minipage}[b]{0.3\linewidth}
\centering
\begin{tabular}{lp{2.5cm}}
       \textbf{Discrete Models} & \textbf{MNIST} \\
       \hline
       \hline
       \textcolor{gg}{\textbf{IGR-I}}        & -57.28 $\pm$ 0.07\\
       \textcolor{rr}{\textbf{IGR-Planar}}   & -56.61 $\pm$ 0.13\\
       \textcolor{bb}{\textbf{IGR-SB}}       & \textbf{-45.12 $\pm$ 1.61} \\
       \textcolor{pp}{\textbf{GS}}           & -59.31 $\pm$ 0.21\\[\smallskipamount]
     \end{tabular}
    \caption{Test log-likelihood on MNIST. Higher is better.}\label{ap:sop}
    \vspace*{9mm}
\end{minipage}\hfill
\begin{minipage}[b]{0.6\linewidth}
\centering
\includegraphics[scale=0.35]{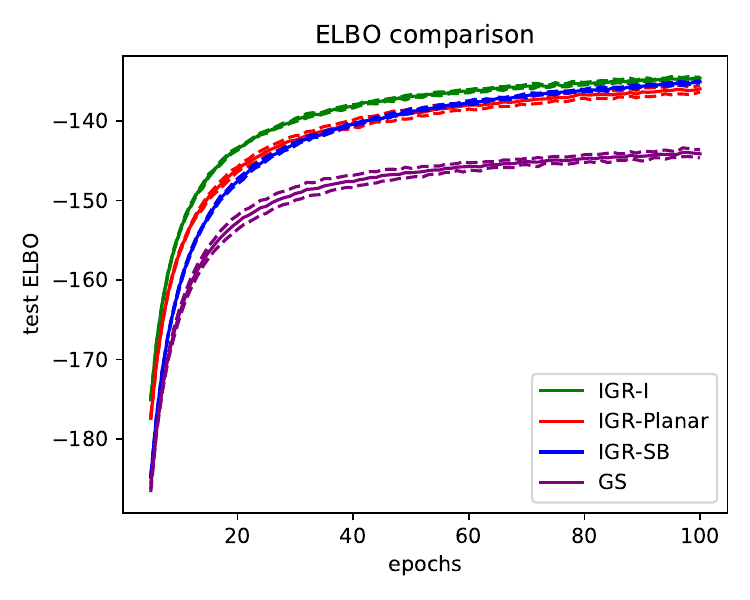}
\captionof{figure}{Test ELBO (of continuous model) on MNIST. Higher is better.
This result is expected to defer from the discrete log-likelihood in Table \ref{rest_cat} (see appendix for details).}
\label{mnist_cat}
\end{minipage}
\end{table}

\subsection{Nonparametric Modeling}

In our last experiment, we show that the truncation strategy that we use in equation \ref{truncation} not only enables the use of continuous relaxations in the countably infinite setting, but that it also allows us to have reparameterizable distributions on $\mathcal{S}^\infty$ (the difference being that one should concentrate most of its mass around the vertices while the other one not necessarily). In order to show this, we follow \citet{nalisnick2016approximate}, who use a VAE with a Dirichlet Process mixture of Gaussians as the prior. For the approximate posterior, they apply the stick-breaking procedure to $K$ Kumaraswamy random variables, where $K$ has to be specified in advance and is thus treated as a hyperparameter \citep{nalisnick2017stick}. We note that this prespecification of $K$ is problematic, as while the prior remains nonparametric, the resulting optimization objective matches the ELBO that would be obtained using a Dirichlet mixture of Gaussians with $K$ components as the prior, effectively losing the nonparametric aspect of the model. In contrast, we use an IGR distribution as the approximate posterior, and the truncation strategy from equation \ref{truncation} allows us to retain the true nonparametric nature of the model.

Since this task does not require a continuous relaxation but just a reparameterizable distribution on the simplex, we use equation \ref{sb_proj} but replace $g$ with the identity function, thus dropping the temperature hyperparameter. We use the label IGR-SB-MM to make this explicit, and use the label DLDPMM for the model of \citet{nalisnick2016approximate}. Table \ref{np} compares the two methods, where we trained DLDPMM with $K=7,9,11,13,15,17$ and report the best result. We can see that not only does IGR enable truly nonparametric inference thus not requiring an expensive hyperparameter search over $K$, but also that this does not come at the cost of decreased performance. We also note that a single IGR-SB-MM run takes the same amount of time as a single DLDPMM run.

 \begin{table}[t]
 \begin{center}
 \scalebox{0.9}{
 \begin{tabular}{lp{2.5cm}p{2.5cm}p{2.5cm}}
 \textbf{Discrete Models} & \textbf{MNIST} & \textbf{Omniglot} \\
 \hline
 \hline
 \textcolor{bb}{\textbf{IGR-SB-MM + SL}} & \textbf{-81.79 $\pm$ 0.35} & \textbf{-129.06 $\pm$ 0.30}\\
 \textcolor{oo}{\textbf{DLDPMM + SL}} & \textbf{-81.90 $\pm$ 0.30} & \textbf{-128.72 $\pm$ 0.22}\\
 [\smallskipamount]
 \end{tabular}}
 \caption{Test log-likelihood on MNIST, FMNIST and Omniglot for IGR and GS. Higher is better.}
 \label{np}
 \end{center}
 \end{table}

\section{Conclusion} \label{conclusion}

In this paper we propose IGR, a flexible discrete reparameterization as an alternative to the GS in
which Gaussian noise is transformed through an invertible function onto the simplex. At the cost of
losing the parameter interpretability of the GS, our method results in a more natural and more flexible
distribution, which has the further advantage of admitting closed form $\mathbb{KL}$ evaluation. We
show that IGR significantly outperforms the GS and that, perhaps surprisingly, this improvement is
not due to this nice theoretical property. Finally, IGR also extends the reparameterization trick to
discrete distributions with countably infinite support and can be incorporated in nonparametric settings.

\section*{Broader Impact}

We do not foresee our work having any negative ethical implications or societal consequences.

\begin{ack}
  We thank Harry Braviner for useful comments. We thank Edmond Cunningham for noticing an error in our Jacobian determinant derivation, which we have now corrected, and also for providing a succinct procedure for deriving it, which we have used. We also thank the Simons Foundation, Sloan Foundation, McKnight Endowment Fund, NIH NINDS 5R01NS100066, NSF 1707398, and the Gatsby Charitable Foundation for support.
\end{ack}

\nocite{steen1969gaussian}
\nocite{kool2020estimating}
\nocite{mnih2016vimco}
\bibliographystyle{abbrvnat}
\bibliography{igr.bib}{}

\begin{thebibliography}{34}
\providecommand{\natexlab}[1]{#1}
\providecommand{\url}[1]{\texttt{#1}}
\expandafter\ifx\csname urlstyle\endcsname\relax
  \providecommand{\doi}[1]{doi: #1}\else
  \providecommand{\doi}{doi: \begingroup \urlstyle{rm}\Url}\fi

\bibitem[Balog et~al.(2017)Balog, Tripuraneni, Ghahramani, and
  Weller]{balog2017lost}
M.~Balog, N.~Tripuraneni, Z.~Ghahramani, and A.~Weller.
\newblock Lost relatives of the gumbel trick.
\newblock In \emph{Proceedings of the 34th International Conference on Machine
  Learning-Volume 70}, pages 371--379. JMLR. org, 2017.

\bibitem[Bengio et~al.(2013)Bengio, L{\'e}onard, and
  Courville]{bengio2013estimating}
Y.~Bengio, N.~L{\'e}onard, and A.~Courville.
\newblock Estimating or propagating gradients through stochastic neurons for
  conditional computation.
\newblock \emph{arXiv preprint arXiv:1308.3432}, 2013.

\bibitem[Bottou(2012)]{bottou2012stochastic}
L.~Bottou.
\newblock Stochastic gradient descent tricks.
\newblock In \emph{Neural networks: Tricks of the trade}, pages 421--436.
  Springer, 2012.

\bibitem[Burda et~al.(2015)Burda, Grosse, and
  Salakhutdinov]{burda2015importance}
Y.~Burda, R.~Grosse, and R.~Salakhutdinov.
\newblock Importance weighted autoencoders.
\newblock \emph{arXiv preprint arXiv:1509.00519}, 2015.

\bibitem[Dinh et~al.(2017)Dinh, Sohl-Dickstein, and Bengio]{dinh2017density}
L.~Dinh, J.~Sohl-Dickstein, and S.~Bengio.
\newblock Density estimation using real nvp.
\newblock \emph{ICLR}, 2017.

\bibitem[Dupont(2018)]{dupont2018learning}
E.~Dupont.
\newblock Learning disentangled joint continuous and discrete representations.
\newblock In \emph{Advances in Neural Information Processing Systems}, pages
  710--720, 2018.

\bibitem[Ferguson(1973)]{ferguson1973bayesian}
T.~S. Ferguson.
\newblock A bayesian analysis of some nonparametric problems.
\newblock \emph{The annals of statistics}, pages 209--230, 1973.

\bibitem[Glynn(1990)]{glynn1990likelihood}
P.~W. Glynn.
\newblock Likelihood ratio gradient estimation for stochastic systems.
\newblock \emph{Communications of the ACM}, 33\penalty0 (10):\penalty0 75--84,
  1990.

\bibitem[Gordon-Rodriguez et~al.(2020)Gordon-Rodriguez, Loaiza-Ganem, and
  Cunningham]{gordon2020continuous}
E.~Gordon-Rodriguez, G.~Loaiza-Ganem, and J.~P. Cunningham.
\newblock The continuous categorical: a novel simplex-valued exponential
  family.
\newblock In \emph{International Conference on Machine Learning}, 2020.

\bibitem[Grathwohl et~al.(2018)Grathwohl, Choi, Wu, Roeder, and
  Duvenaud]{grathwohl2017backpropagation}
W.~Grathwohl, D.~Choi, Y.~Wu, G.~Roeder, and D.~Duvenaud.
\newblock Backpropagation through the void: Optimizing control variates for
  black-box gradient estimation.
\newblock \emph{ICLR}, 2018.

\bibitem[Gu et~al.(2015)Gu, Levine, Sutskever, and Mnih]{gu2015muprop}
S.~Gu, S.~Levine, I.~Sutskever, and A.~Mnih.
\newblock Muprop: Unbiased backpropagation for stochastic neural networks.
\newblock \emph{arXiv preprint arXiv:1511.05176}, 2015.

\bibitem[Jang et~al.(2017)Jang, Gu, and Poole]{jang2017categorical}
E.~Jang, S.~Gu, and B.~Poole.
\newblock Categorical reparameterization with gumbel-softmax.
\newblock \emph{ICLR}, 2017.

\bibitem[Johnson et~al.(2016)Johnson, Duvenaud, Wiltschko, Adams, and
  Datta]{johnson2016composing}
M.~Johnson, D.~K. Duvenaud, A.~Wiltschko, R.~P. Adams, and S.~R. Datta.
\newblock Composing graphical models with neural networks for structured
  representations and fast inference.
\newblock In \emph{Advances in neural information processing systems}, pages
  2946--2954, 2016.

\bibitem[Kingma and Welling(2014)]{kingma2014auto}
D.~P. Kingma and M.~Welling.
\newblock Auto-encoding variational bayes.
\newblock \emph{ICLR}, 2014.

\bibitem[Kingma et~al.(2016)Kingma, Salimans, Jozefowicz, Chen, Sutskever, and
  Welling]{kingma2016improved}
D.~P. Kingma, T.~Salimans, R.~Jozefowicz, X.~Chen, I.~Sutskever, and
  M.~Welling.
\newblock Improved variational inference with inverse autoregressive flow.
\newblock In \emph{Advances in neural information processing systems}, pages
  4743--4751, 2016.

\bibitem[Kool et~al.(2020)Kool, van Hoof, and Welling]{kool2020estimating}
W.~Kool, H.~van Hoof, and M.~Welling.
\newblock Estimating gradients for discrete random variables by sampling
  without replacement.
\newblock In \emph{International Conference on Learning Representations}, 2020.
\newblock URL \url{https://openreview.net/forum?id=rklEj2EFvB}.

\bibitem[Kusner and Hern{\'a}ndez-Lobato(2016)]{kusner2016gans}
M.~J. Kusner and J.~M. Hern{\'a}ndez-Lobato.
\newblock Gans for sequences of discrete elements with the gumbel-softmax
  distribution.
\newblock \emph{arXiv preprint arXiv:1611.04051}, 2016.

\bibitem[Linderman et~al.(2018)Linderman, Mena, Cooper, Paninski, and
  Cunningham]{linderman2018reparameterizing}
S.~Linderman, G.~Mena, H.~Cooper, L.~Paninski, and J.~Cunningham.
\newblock Reparameterizing the birkhoff polytope for variational permutation
  inference.
\newblock In \emph{International Conference on Artificial Intelligence and
  Statistics}, pages 1618--1627, 2018.

\bibitem[Maddison et~al.(2017)Maddison, Mnih, and Teh]{maddison2017concrete}
C.~J. Maddison, A.~Mnih, and Y.~W. Teh.
\newblock The concrete distribution: A continuous relaxation of discrete random
  variables.
\newblock \emph{ICLR}, 2017.

\bibitem[Miller et~al.(2017)Miller, Foti, D'Amour, and
  Adams]{miller2017reducing}
A.~Miller, N.~Foti, A.~D'Amour, and R.~P. Adams.
\newblock Reducing reparameterization gradient variance.
\newblock In \emph{Advances in Neural Information Processing Systems}, pages
  3708--3718, 2017.

\bibitem[Mnih and Rezende(2016{\natexlab{a}})]{mnih2016variational}
A.~Mnih and D.~Rezende.
\newblock Variational inference for monte carlo objectives.
\newblock In \emph{International Conference on Machine Learning}, pages
  2188--2196, 2016{\natexlab{a}}.

\bibitem[Mnih and Rezende(2016{\natexlab{b}})]{mnih2016vimco}
A.~Mnih and D.~J. Rezende.
\newblock Variational inference for monte carlo objectives.
\newblock \emph{CoRR}, 1602.06725, 2016{\natexlab{b}}.
\newblock URL \url{http://arxiv.org/abs/1602.06725}.

\bibitem[Nalisnick and Smyth(2017)]{nalisnick2017stick}
E.~Nalisnick and P.~Smyth.
\newblock Stick-breaking variational autoencoders.
\newblock In \emph{ICLR}, 2017.

\bibitem[Nalisnick et~al.(2016)Nalisnick, Hertel, and
  Smyth]{nalisnick2016approximate}
E.~Nalisnick, L.~Hertel, and P.~Smyth.
\newblock Approximate inference for deep latent gaussian mixtures.
\newblock In \emph{NIPS Workshop on Bayesian Deep Learning}, volume~2, 2016.

\bibitem[Raiko et~al.(2014)Raiko, Berglund, Alain, and
  Dinh]{raiko2014techniques}
T.~Raiko, M.~Berglund, G.~Alain, and L.~Dinh.
\newblock Techniques for learning binary stochastic feedforward neural
  networks, 2014.

\bibitem[Rezende and Mohamed(2015)]{rezende2015variational}
D.~Rezende and S.~Mohamed.
\newblock Variational inference with normalizing flows.
\newblock In \emph{International Conference on Machine Learning}, pages
  1530--1538, 2015.

\bibitem[Rezende et~al.(2014)Rezende, Mohamed, and
  Wierstra]{rezende2014stochastic}
D.~J. Rezende, S.~Mohamed, and D.~Wierstra.
\newblock Stochastic backpropagation and approximate inference in deep
  generative models.
\newblock In \emph{International Conference on Machine Learning}, pages
  1278--1286, 2014.

\bibitem[Robbins and Monro(1951)]{robbins1951stochastic}
H.~Robbins and S.~Monro.
\newblock A stochastic approximation method.
\newblock \emph{The annals of mathematical statistics}, pages 400--407, 1951.

\bibitem[Roeder et~al.(2017)Roeder, Wu, and Duvenaud]{roeder2017sticking}
G.~Roeder, Y.~Wu, and D.~K. Duvenaud.
\newblock Sticking the landing: Simple, lower-variance gradient estimators for
  variational inference.
\newblock In \emph{Advances in Neural Information Processing Systems}, pages
  6925--6934, 2017.

\bibitem[Steen et~al.(1969)Steen, Byrne, and Gelbard]{steen1969gaussian}
N.~Steen, G.~Byrne, and E.~Gelbard.
\newblock Gaussian quadratures for the integrals $\int_0^{\infty} \exp(-x^2)
  f(x)dx$ and $\int_0^b \exp(-x^2) f(x)dx$.
\newblock \emph{Mathematics of Computation}, pages 661--671, 1969.

\bibitem[Stirn et~al.(2019)Stirn, Jebara, and Knowles]{stirn2019new}
A.~Stirn, T.~Jebara, and D.~Knowles.
\newblock A new distribution on the simplex with auto-encoding applications.
\newblock In \emph{Advances in Neural Information Processing Systems}, pages
  13670--13680, 2019.

\bibitem[Tucker et~al.(2017)Tucker, Mnih, Maddison, Lawson, and
  Sohl-Dickstein]{tucker2017rebar}
G.~Tucker, A.~Mnih, C.~J. Maddison, J.~Lawson, and J.~Sohl-Dickstein.
\newblock Rebar: Low-variance, unbiased gradient estimates for discrete latent
  variable models.
\newblock In \emph{Advances in Neural Information Processing Systems}, pages
  2627--2636, 2017.

\bibitem[Williams(1992)]{williams1992simple}
R.~J. Williams.
\newblock Simple statistical gradient-following algorithms for connectionist
  reinforcement learning.
\newblock \emph{Machine learning}, 8\penalty0 (3-4):\penalty0 229--256, 1992.

\bibitem[Xie and Ermon(2019)]{xie2019reparameterizable}
S.~M. Xie and S.~Ermon.
\newblock Reparameterizable subset sampling via continuous relaxations.
\newblock In \emph{International Joint Conference on Artificial Intelligence},
  2019.

\end{thebibliography}

\appendix
\newpage
\hrule height 4pt
\vskip 0.25in
\vskip -\parskip
\begin{center}
{\LARGE\bf Appendix for: Invertible Gaussian Reparameterization}
\end{center}

\vskip 0.29in
\vskip -\parskip
\hrule height 1pt
\vskip 0.09in

\section{Computing the determinant of the Jacobian of the $\text{softmax}_{++}$}

As mentioned in section 3.1, we can use the matrix determinant lemma to efficiently compute the determinant of the Jacobian of the $\text{softmax}_{++}$. 
If, for $i=1,\dots,K-1$, we let:
\begin{equation}
    \begin{split}
      g_{i} 
      := g\left(y, \tau, \delta\right)_{i} 
      := \frac{\exp\left(y_{i} / \tau \right)}{\sum_{k=1}^{K-1} \exp\left(y_{k} / \tau\right) + \delta}
    \end{split}
\end{equation}
then it is not hard to see that when $i \neq j$ then:
\begin{equation}
    \begin{split}
      \frac{\partial g_{i}}{\partial y_{j}}
      &= -\frac{\exp\left(y_{i} / \tau \right)}{\left(\sum_{k=1}^{K-1} \exp\left(y_{k} / \tau\right) + \delta\right)^{2}}
      \left(\frac{1}{\tau} \exp\left(y_{j} / \tau\right)\right)
      \\
      &=
      - \frac{1}{\tau} g_{i} g_{j}
    \end{split}
\end{equation}
and that:
\begin{equation}
    \begin{split}
      \frac{\partial g_{i}}{\partial y_{i}}
      &=
      \left(\frac{1}{\tau}\right)
      \left(\frac{\exp\left(y_{i} / \tau \right)}{\sum_{k=1}^{K-1} \exp\left(y_{k} / \tau\right) + \delta}\right)
      \\
      &-
      \frac{\exp\left(y_{i} / \tau \right)}{\left(\sum_{k=1}^{K-1} \exp\left(y_{k} / \tau\right) + \delta\right)^{2}}
      \left(\frac{1}{\tau} \exp\left(y_{i} / \tau\right)\right)
      \\
      &= \frac{1}{\tau} g_{i} \left(1 - g_{i}\right).
    \end{split}
\end{equation}

Combining the previous equations we have that:
\begin{equation}
  \begin{split}
  J_g(y, \tau) 
    & =  
  \frac{1}{\tau}
  \begin{pmatrix} 
    g_{1}\left(1 - g_{1}\right) & 
    - g_{1} g_{2} & 
    \dots & 
    - g_{1} g_{K - 1}
    \\ 
    - g_{2} g_{1} & 
    g_{2} \left(1 - g_{2}\right)& 
    \dots & 
    - g_{2} g_{K - 1}
    \\ 
    \vdots & 
    \vdots & 
    \ddots &
    \vdots 
    \\ 
    - g_{K - 1} g_{1} & 
    - g_{K - 1} g_{2} & 
    \dots & 
    g_{K - 1} \left(1 - g_{K - 1}\right)
  \end{pmatrix}
    \\
    & =
    \frac{1}{\tau} \left(\text{diag}\left(g\right) - g g^{T}\right).
  \end{split}
\end{equation}
Given the previous expression, we can compute the determinant as follows:
\begin{equation}
    \begin{split}
      \left| \det J_{g}\left(y, \tau\right)\right|
      &=
      \left| \frac{1}{\tau} \left(\text{diag}\left(g\right) - g g^{T}\right)\right|
      \\
      &=
      \frac{1}{\tau^{K - 1}} \left|\text{diag}\left(g\right)\right| \left(1 - g^{T} \text{diag}\left(g\right)^{-1} g\right)
      \\
      &=
      \frac{1}{\tau^{K - 1}} \left(\prod_{i=1}^{K - 1} g_{i}\right) \left(1 - \sum_{j=1}^{K - 1} g_{j}\right).
    \end{split}
\end{equation}
From the previous expression the log determinant becomes apparent:
\begin{equation*}
    \begin{split}
      \log \left| \det J_{g}\left(y, \tau\right)\right|
      =
      -\left(K - 1\right) \log \tau + \sum_{i=1}^{K - 1} \log g_{i} + \log\left(1 - \sum_{j=1}^{K - 1} g_{j}\right).
    \end{split}
\end{equation*}


\section{Proofs of propositions}

\textbf{Proposition 1}: For any $\delta > 0$, the following holds:
\begin{equation}
\lim_{\tau \rightarrow 0} \text{softmax}_{++}(y / \tau) = h(y) := \begin{cases} e_{k^*}, \text{ if
}k^*=\displaystyle \argmax_{k=1,\dots,K-1}(y_k)\text{ and }\max_{k=1,\dots,K-1}(y_k) > 0\\ 0,
\hspace{3pt}\text{ if }\displaystyle \max_{k=1,\dots,K-1}(y_k) < 0
\end{cases}
\end{equation}
where $e_k \in \mathbb{R}^{K-1}$ is the one-hot vector with a $1$ in its $k$-th coordinate.\\

\textbf{Proof}: We will assume that $\argmax_{k=1,\dots,K-1}(y_k)$ is unique, and denote $y^* =
\max_{k=1,\dots,K-1}(y_k)$. We then have: \begin{equation} \text{softmax}_{++}(y/\tau)_k =
\dfrac{e^{y_k / \tau}}{ \sum_{j=1}^{K-1} e^{y_j / \tau} + \delta} = \dfrac{e^{(y_k - y^*) / \tau}}{
\sum_{j=1}^{K-1} e^{(y_j-y^*) / \tau} + \delta e^{-y^*/\tau}} \end{equation} If $y_k < y^*$, the
numerator goes to $0$ as $\tau \rightarrow 0$, while it is equal to $1$ if $y_k = y^*$. The
denominator goes to $\infty$ if $y^* < 0$, while it goes to $1$ if $y^* > 0$. Combining these
observations finishes the proof.  \qed\\

\textbf{Proposition 2}: If $y_k \sim \mathcal{N}(\mu_k, \sigma_k)$ for $k=1,\dots, K-1$, and we
define the discrete random variable $H$ by $H = k$ if $h(y)=e_k$ and $H = K$ if $h(y)=0$, then:
\begin{equation}\label{discrete_recovered} \mathbb{P}(H=k)=\begin{cases} \displaystyle
\int_{0}^\infty \phi\left(\dfrac{t - \mu_k}{\sigma_k}\right) \prod_{j \neq k} \Phi\left(\dfrac{t -
\mu_j}{\sigma_j}\right) dt ,\text{ if }k=1,\dots,K-1\\ \displaystyle \prod_{j=1}^{K-1}
\Phi\left(-\dfrac{\mu_j}{\sigma_j}\right) \hspace{87pt}, \text{ if }k=K \end{cases} \end{equation}
where $\phi$ and $\Phi$ are the standard Gaussian pdf and cdf, respectively.\\

\textbf{Proof}: For $k=1,\dots, K-1$, we have:
\begin{align}
\mathbb{P}(H=k) & = \displaystyle \int_{\{y: y_k \geq y_1, \dots, y_k \geq y_{K-1}, y_k \geq
0\}}p(y)dy \\ & = \int_{0}^{\infty}\int_{-\infty}^{y_k}\cdots \int_{-\infty}^{y_k} \prod_{j=1}^{K-1}
\dfrac{1}{\sigma_j}\phi\left(\dfrac{y_j - \mu_j}{\sigma_j}\right) dy_1 \cdots dy_{k-1}dy_{k+1}\cdots
dy_{K-1} dy_k \\ & = \int_0^\infty  \dfrac{1}{\sigma_k}\phi\left(\dfrac{y_k -
\mu_k}{\sigma_k}\right) \prod_{j \neq k} \left(
\int_{-\infty}^{y_k}\dfrac{1}{\sigma_j}\phi\left(\dfrac{y_j - \mu_j}{\sigma_j}\right) dy_j\right)
dy_k \\ & = \int_0^\infty \dfrac{1}{\sigma_k}\phi\left(\dfrac{y_k - \mu_k}{\sigma_k}\right)
\prod_{j\neq k} \Phi\left(\dfrac{y_k - \mu_j}{\sigma_j}\right) dy_k
\end{align}
which finishes the first part of the proof. The remaining probability, $\mathbb{P}(H=K)$ can
obviously be recovered as one minus the sum of the above probabilities, but we can also obtain the
following expression:
\begin{align} \mathbb{P}(H=K) & = \displaystyle \int_{\{y:y_1 \leq 0,\dots, y_{K-1}\leq 0\}}p(y)dy\\
& = \int_{-\infty}^0 \cdots \int_{-\infty}^0 \prod_{j=1}^{K-1}
\dfrac{1}{\sigma_j}\phi\left(\dfrac{y_j - \mu_j}{\sigma_j}\right)dy_1 \cdots dy_{K-1} \\ & =
\prod_{j=1}^{K-1} \Phi\left(-\dfrac{\mu_j}{\sigma_j}\right) \end{align}
which finishes the proof.
\qed

In our experiments with RELAX \cite{grathwohl2017backpropagation}
in section \ref{ap:re} of the appendix we approximate the required
integrals using a Gaussian quadrature as in \citet{steen1969gaussian}, and backpropagate through this
procedure. Note that the involved integrals are one-dimensional and thus can be accurately
approximated with quadrature methods. Although we found better performance
with these approximations than with a Monte Carlo approximation, we found the method prone to numerical instabilities, which we solved by limiting the range of values that $\mu$ and $\sigma$ are allowed take as follows:
\begin{equation}
  \begin{split}
    \mu
    &=
    -5 \text{ tanh}\left(\mu'\right)
    \\
    \sigma
    &= 0.5 + 2 \text{ sigmoid}\left(\sigma'\right)
  \end{split}
\end{equation}
where $\mu'$ and $\sigma'$ are the parameters that we optimize over.

\section{Variational Autoencoders}\label{ap:vaes}


As mentioned in the main manuscript, our VAE experiments closely follow \citet{maddison2017concrete}: we use the same continuous objective and the same evaluation metrics.
The experiments differ to \citet{jang2017categorical} since they use a $\mathbb{KL}$ term as in
equation 8 of the main manuscript, whereas \citet{maddison2017concrete} use a continuous
$\mathbb{KL}$ as in equation 9 of the main manuscript. Using the former $\mathbb{KL}$ results in
optimizing a continuous objective which is not a log-likelihood lower bound anymore, which is mainly
why we followed \citet{maddison2017concrete}.

In addition to the reported comparisons in the main manuscript, we include further comparisons in Table \ref{ta:vae_tr} reporting the discretized training ELBO instead.

\begin{table}[]
 \begin{center}
 \scalebox{0.9}{
 \begin{tabular}{lp{2.5cm}p{2.5cm}p{2.5cm}}
 \textbf{Model} & \textbf{MNIST} & \textbf{FMNIST} & \textbf{Omniglot} \\
 \hline
 \hline
 \textcolor{gg}{\textbf{IGR-I}}
 & -131.34 $\pm$ 0.15         & -67.45 $\pm$ 0.18         & -147.50 $\pm$ 0.18          \\
 \textcolor{rr}{\textbf{IGR-Planar}}
 & -132.74 $\pm$ 0.20         & -67.90 $\pm$ 0.82         & -148.81 $\pm$ 0.17          \\
 \textcolor{bb}{\textbf{IGR-SB}}
 & \textbf{-127.67 $\pm$ 0.23}& \textbf{-63.78 $\pm$ 0.47}& \textbf{-144.47 $\pm$ 0.12} \\
 \vspace{0.05cm}
 \textcolor{pp}{\textbf{GS}}
 & -154.01 $\pm$ 0.53         & -84.99 $\pm$ 0.64         & -159.99 $\pm$ 0.48          \\
 \hline
 \textcolor{gg}{\textbf{IGR-I + SL}}
 & -131.06 $\pm$ 0.26         & -67.12 $\pm$ 0.37         & -146.68 $\pm$ 0.18          \\
 \textcolor{rr}{\textbf{IGR-Planar + SL}}
 & -132.59 $\pm$ 0.18         & -67.46 $\pm$ 0.24         & -148.22 $\pm$ 0.38          \\
 \textcolor{bb}{\textbf{IGR-SB + SL}}
 & \textbf{-127.64 $\pm$ 0.24}& \textbf{-63.42 $\pm$ 0.24}& \textbf{-143.39 $\pm$ 0.19} \\
 \textcolor{pp}{\textbf{GS + SL}}
 & -153.58 $\pm$ 0.53         & -85.99 $\pm$ 1.03         & -159.25 $\pm$ 0.59
 \\[\smallskipamount]
 \end{tabular}}
 \caption{Discretized Train ELBO (not log-likelihood) on MNIST, FMNIST and Omniglot for IGR and GS.
 Higher is better.}
 \label{ta:vae_tr}
 \end{center}
 \end{table}

\section{Other Estimators}\label{ap:re}

\citet{tucker2017rebar} and \citet{grathwohl2017backpropagation} proposed REBAR and RELAX, respectively. These are
variance reduction techniques which heavily lean on the GS
to improve the variance of the obtained
gradients. We make several important notes: First, REBAR is a special case of RELAX, so that we will
only compare against RELAX. Second, RELAX takes advantage of the parameter interpretability of the
GS, as it considers the gradients of the relaxed objective as approximations to the gradients of the
objective of interest:
\begin{equation}\label{GS_grad} \nabla_{\alpha}\mathbb{E}_{z \sim \alpha}[f(z)] \approx
\nabla_{\alpha}\mathbb{E}_{\tilde{q}_{\alpha, \tau}(\tilde{z})}[\tilde{f}(\tilde{z})]
\end{equation}
where $\alpha$ is a discrete distribution, which we think of as a vector of length $K$ and
$\tilde{q}_{\alpha, \tau}$ is a GS distribution. RELAX builds upon equation \ref{GS_grad} to develop
an estimator with reduced variance. Extending this observation to IGR is not immediately
straightforward, as $\nabla_{\mu, \sigma} \mathbb{E}_{q_{\mu, \sigma,
\tau}(\tilde{z})}[\tilde{f}(\tilde{z})]$ is not an approximation to the gradient on the left hand
size of the above equation: it is not even the same shape. However, thanks to proposition 2 we can
parameterize a discrete distribution using $\mu$ and $\sigma$, so that $\alpha(\mu, \sigma)$ is the
discrete distribution given by proposition 2. This way, instead of directly optimizing over the
discrete distribution, we optimize over its parameters, $\mu$ and $\sigma$, so that the gradient of
interest becomes $\nabla_{\mu, \sigma} \mathbb{E}_{z \sim \alpha(\mu, \sigma)}[f(z)]$, and its
corresponding approximation:
\begin{equation}\label{IGR_grad} \nabla_{\mu, \sigma} \mathbb{E}_{z \sim \alpha(\mu, \sigma)}[f(z)]
\approx \nabla_{\mu, \sigma} \mathbb{E}_{\tilde{q}_{\mu, \sigma,
\tau}(\tilde{z})}[\tilde{f}(\tilde{z})] \end{equation}
where $\tilde{q}_{\mu, \sigma, \tau}$ is an IGR distribution, thus enabling the use of RELAX along
IGR. Third, it should also be noted that the bias and variance of the gradient estimator of RELAX
are central points of discussion by \citet{grathwohl2017backpropagation}. However, comparing bias and variance
between the GS and IGR is a difficult task, as they are intrinsically approximating different
gradients (equations \ref{GS_grad} and \ref{IGR_grad}, respectively). To make the fairest possible
comparison, we compare between IGR and the GS not by trying to estimate biases and variances, but by
empirically comparing the recovered discrete objectives. Ultimately, bias and variance of a
stochastic gradient estimator are used as proxies for how adequately optimized the corresponding
objective will be, so that directly comparing on this metric is sensible. We show results of running
IGR and GS with and without RELAX in Table \ref{ap:table_relax}.

\begin{table}[h]
\centering
\begin{tabular}{lp{2.5cm}}
       \textbf{Discrete Models} & \textbf{MNIST} \\
       \hline
       \hline
       \textcolor{gg}{\textbf{IGR-I}}                & -94.18  $\pm$ 0.37          \\
       \textcolor{pp}{\textbf{GS}}                   & -103.80 $\pm$ 0.73          \\
       \textcolor{gg}{\textbf{IGR-I + RELAX}}        & \textbf{-82.03 $\pm$ 2.18}  \\
       \textcolor{pp}{\textbf{GS + RELAX}}           & \textbf{-83.41  $\pm$ 1.11} \\[\smallskipamount]
     \end{tabular}
    \caption{Test log-likelihood on MNIST for nonlinear architecture. Higher is better.}\label{ap:table_relax}
\end{table}

Finally, \citet{kool2020estimating} proposed USPGBL, an unbiased estimator (unlike the GS or IGR, which are biased), which is based on sampling without replacement. Their method requires using several approximate posterior samples to estimate the ELBO. We used $S=4$ samples, and for a fair comparison against GS and IGR, we also estimated the ELBO using $4$ samples (instead of $1$, which we used in every other experiment). Results are in Table \ref{ap:wo} and we can see that again, IGR performs best.

\begin{table}[h]
\centering
\begin{tabular}{lp{2.5cm}p{3.5cm}}
       \textbf{Discrete Models} & \textbf{MNIST ($S=4$)}\\
       \hline
       \hline
       \textcolor{gg}{\textbf{IGR-I}} & -118.45 $\pm$ 3.02                  \\
       \textcolor{pp}{\textbf{GS}} & -126.84 $\pm$ 2.20                     \\
       \textcolor{gg}{\textbf{IGR-I + RELAX}} & \textbf{-100.58 $\pm$ 3.38} \\
       \textcolor{pp}{\textbf{GS + RELAX}} & -112.76  $\pm$ 1.72            \\
       \textcolor{oo}{\textbf{USPGBL}} & -106.89 $\pm$ 1.37                 \\
     \end{tabular}
    \caption{Test ELBO on MNIST for nonlinear architecture with 4
    samples. Higher is better.}\label{ap:wo}
\end{table}

\section{Architecture and hyperparameter details} \label{ap:archi}

In this section we describe the hyperparameters and architecture for the VAEs we use. The
choice of hyperparameters and architecture are aligned with \cite{maddison2017concrete,
jang2017categorical, tucker2017rebar, grathwohl2017backpropagation, kool2020estimating}.

 \begin{itemize}
   \item \textbf{Linear Architecture: 784 - 200 - 784}
   \begin{itemize}
    \item Encoder: One fully connected dense layers of 200 units with linear activation.
    \item Decoder: Symmetrical to the Encoder. One fully connected dense layer of 200 units with
    linear activation.
   \end{itemize}
   \item \textbf{Non-linear Architecture: 784 $\sim$ 512 $\sim$ 256 - 200 - 256 $\sim$
   512 $\sim$ 784}
   \begin{itemize}
    \item Encoder: Two fully connected dense layers of 512 units and 256 units respectively. The
    nonlinear activations are ReLu.

    \item Decoder: Symmetrical to the Encoder. Two fully connected dense layers of 256 units and 512
    units respectively. The nonlinear activations are ReLu.
   \end{itemize}
 \end{itemize}

The hyperparameters are shared across the models. The only thing that changes is the temperature,
which is selected through cross validation as specified in the main manuscript. We use the following
configuration:
 \begin{itemize}
 \item Batch size = 100
 \item Epochs = 300 - 500
 \item Learning Rate $\in \{1.e-4, 3.e-4\}$
 \item Adam with $\beta_{1}=0.9, \beta_{2}=0.999$
 \item Categories = 10
 \item Number of Discrete Variables = 20
 \end{itemize}

For the \textbf{structure output prediction} task the architecture used is:
\begin{itemize}
  \item \textbf{Four-layered non-linear architecture 240 $\sim$ 240 - (first sample) \&
  240 $\sim$ 240 - (second sample)}
  \begin{itemize}
    \item First sample is taken from double-layer 240 with tahn activation followed by a
    layer with 240 units with a linear activation
    \item Second sample is taken as above
  \end{itemize}
\end{itemize}

The hyperparameters used are
 \begin{itemize}
 \item Batch size = 100
 \item Epochs = 100
 \item Learning Rate = 1.e-3
 \item Weight Decay = 1.e-3
 \item Adam with $\beta_{1}=0.9, \beta_{2}=0.999$
 \end{itemize}

For the \textbf{nonparameteric mixture model} the architecture used is:
\begin{itemize}
  \item \textbf{Nonlinear architecture 784 $\sim$ 200 $\sim$ 200 - 200 - 200 $\sim$ 200 - 784}
  \begin{itemize}
    \item Encoder: 3 fully connected dense layers with 200 units and a ReLu activation.
    \item Decoder: 3 fully connected dense layers with 200 units and a ReLu activation.
  \end{itemize}
\end{itemize}
The hyperparameters used are
 \begin{itemize}
 \item Batch size = 100
 \item Epochs = 300
 \item Learning Rate = 3.e-4
 \item Continuous Dimensionality = 50
 \item Max number of mixtures = 20 (not necessarily used)
 \item Adam with $\beta_{1}=0.9, \beta_{2}=0.999$
 \end{itemize}

\newpage

\section{Approximating Discrete Distributions} \label{approx_disc}
 Next we compare the GS and the IGR in approximating discrete distributions. We took 1,000 samples
 of the learned parameters of the IGR from solving equation 14 from the main manuscript.

 \begin{figure}[H]
  \centering
  \includegraphics[width=0.45\textwidth,height=4cm]{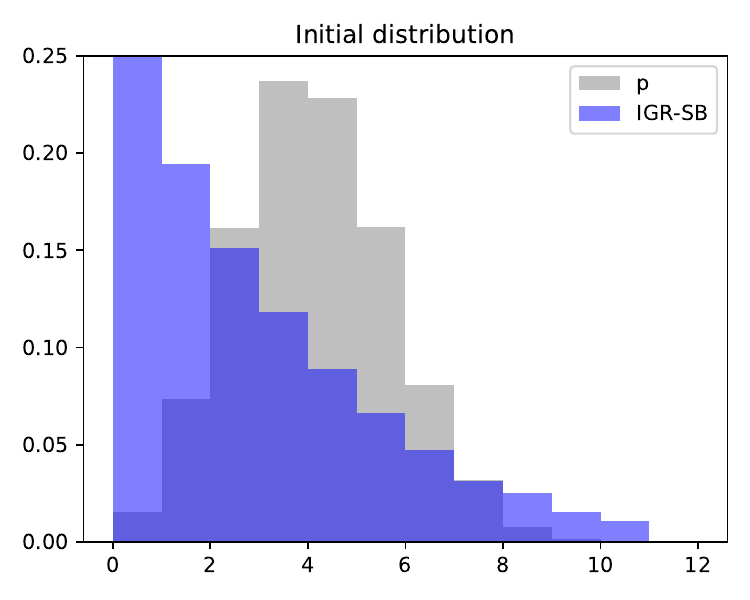}
  \includegraphics[width=0.45\textwidth,height=4cm]{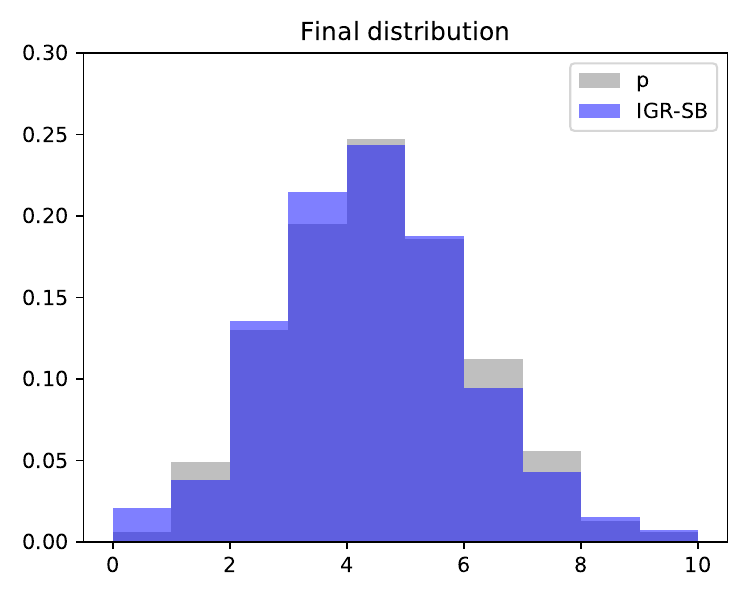}
  \caption{IGR approximation to a Binomial$(N=12, p=0.3)$}\label{fig:binomial}
 \end{figure}

 \begin{figure}[H]
  \centering
  \includegraphics[width=0.45\textwidth,height=4cm]{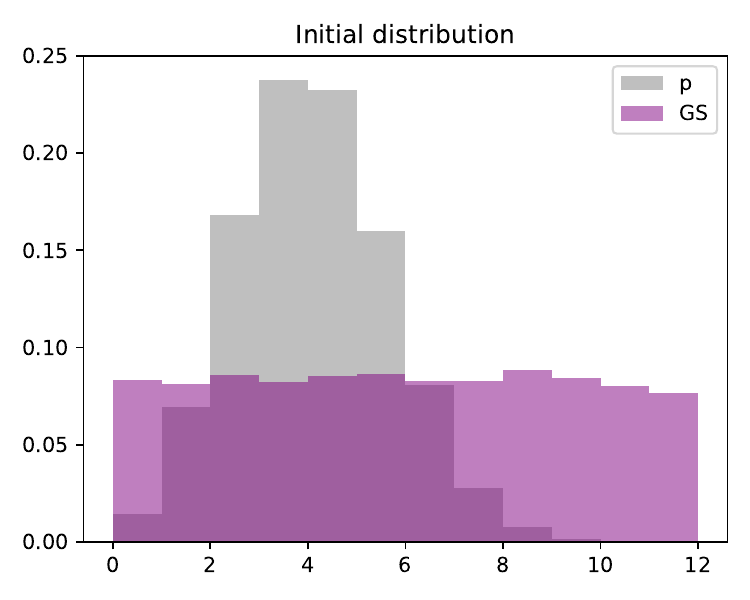}
  \includegraphics[width=0.45\textwidth,height=4cm]{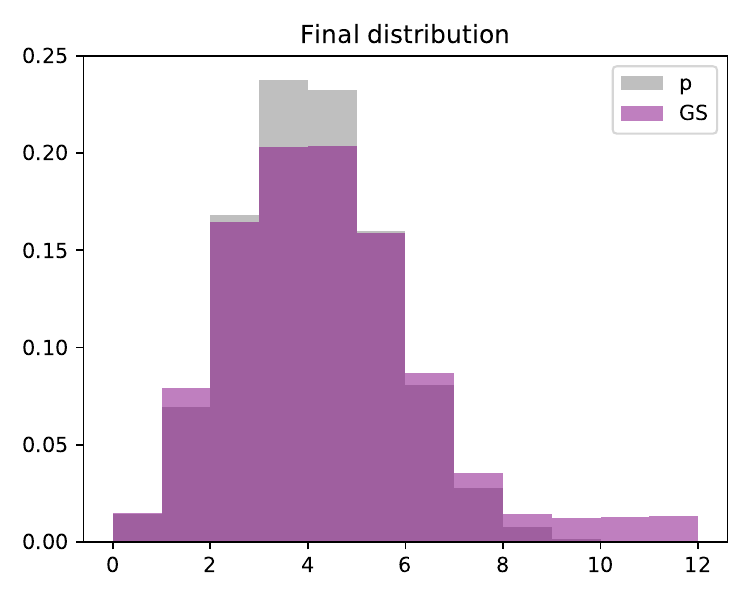}
  \caption{GS approximation to a Binomial$(N=12, p=0.3)$}\label{fig:binomial_gs}
 \end{figure}

We observe how both methods approximate the Binomial adequately, although it seems that the
advantage of the IGR-SB to better approximate countably infinite distributions was not translated to
this simple example.

 \begin{figure}[H]
  \centering
  \includegraphics[width=0.45\textwidth,height=4cm]{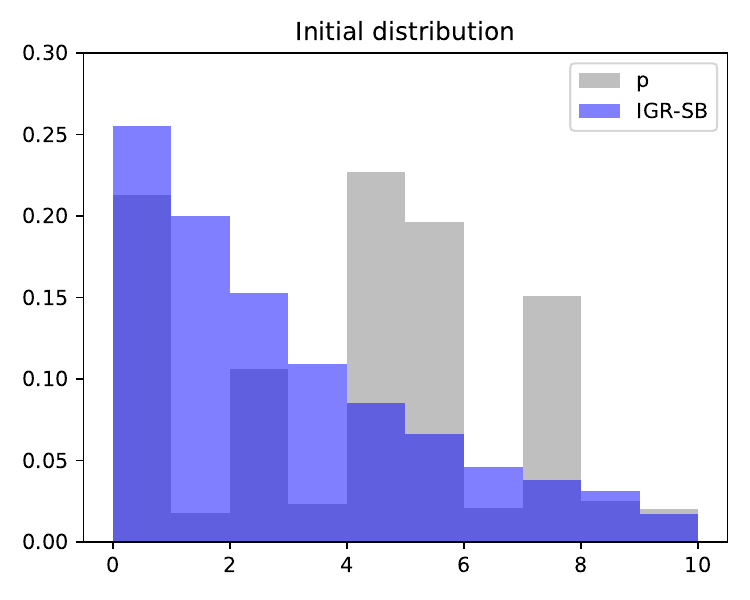}
  \includegraphics[width=0.45\textwidth,height=4cm]{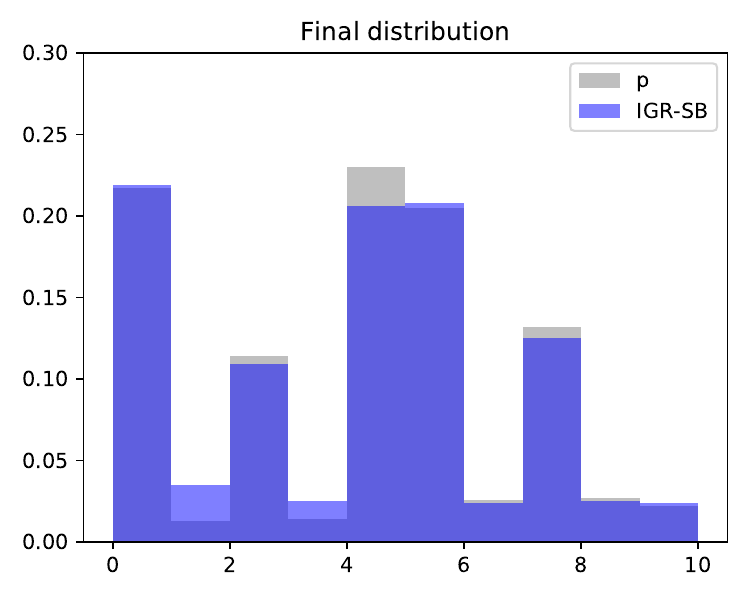}
  \caption{IGR approximation to a Discrete defined as $p=(\frac{10}{46}, \frac{1}{46},
  \frac{5}{46}, \frac{1}{46}, \frac{10}{46}, \frac{10}{46}, \frac{1}{46}, \frac{6}{46},
  \frac{1}{46}, \frac{1}{46})$.}\label{fig:discrete}
 \end{figure}

 \begin{figure}[H]
  \centering
  \includegraphics[width=0.45\textwidth,height=4cm]{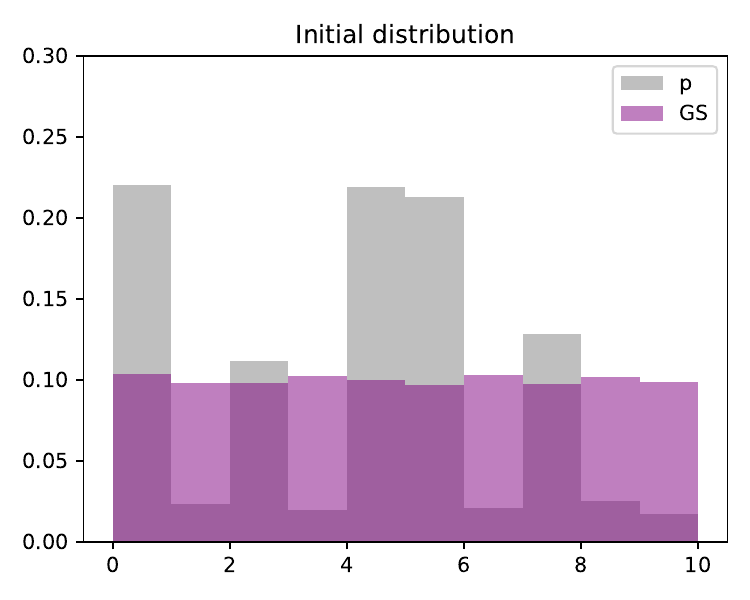}
  \includegraphics[width=0.45\textwidth,height=4cm]{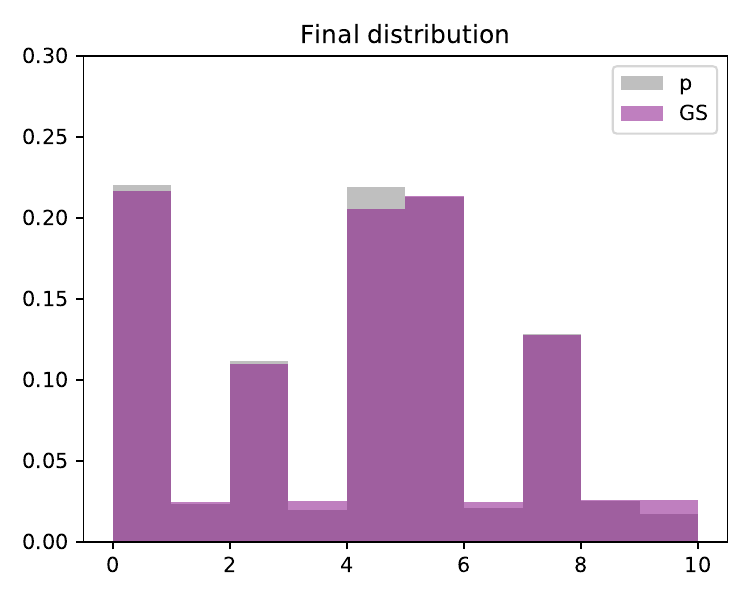}
  \caption{GS approximation to a Discrete defined as $p=(\frac{10}{46}, \frac{1}{46},
  \frac{5}{46}, \frac{1}{46}, \frac{10}{46}, \frac{10}{46}, \frac{1}{46}, \frac{6}{46},
  \frac{1}{46}, \frac{1}{46})$.}\label{fig:discrete}
 \end{figure}

Results for this discrete distribution are similar to those observed on the Binomial.

 \begin{figure}[H]
  \centering
  \includegraphics[width=0.45\textwidth,height=4cm]{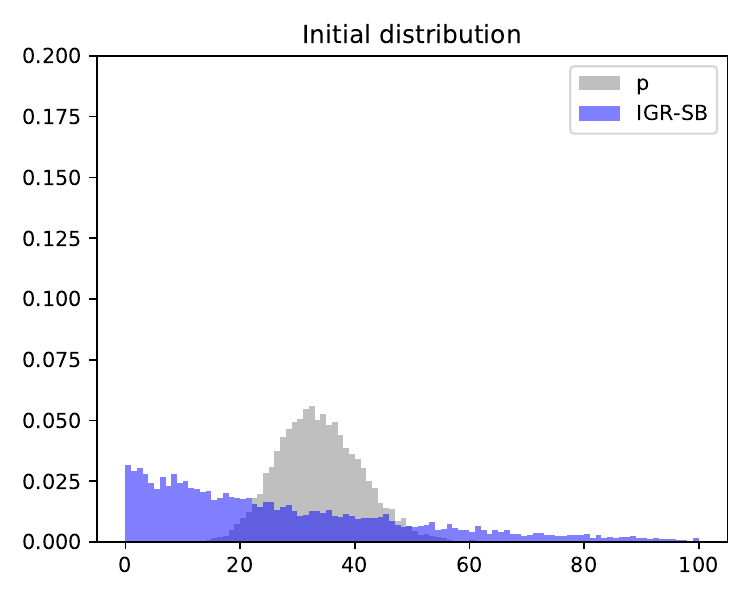}
  \includegraphics[width=0.45\textwidth,height=4cm]{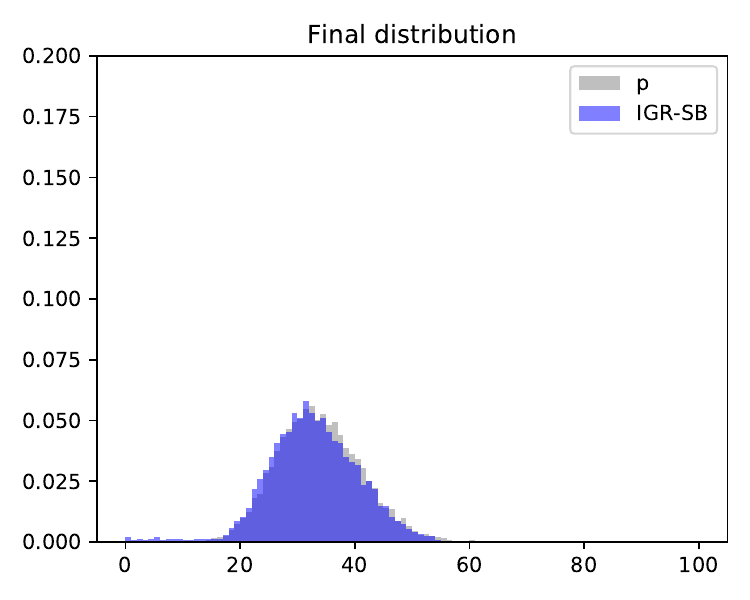}
  \caption{IGR-SB approximation to a Negative Binomial$(r=50,
  p=0.6)$.}\label{fig:negative_binomial}
 \end{figure}

 \begin{figure}[H]
  \centering
  \includegraphics[width=0.45\textwidth,height=4cm]{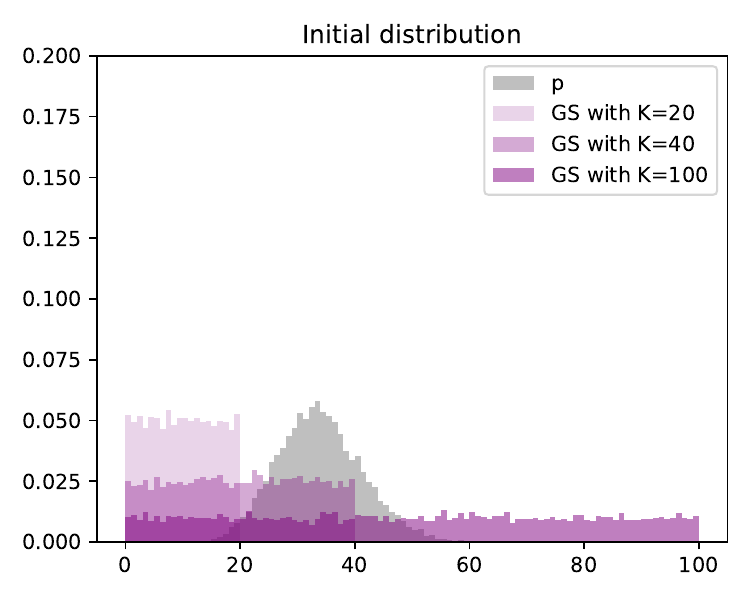}
  \includegraphics[width=0.45\textwidth,height=4cm]{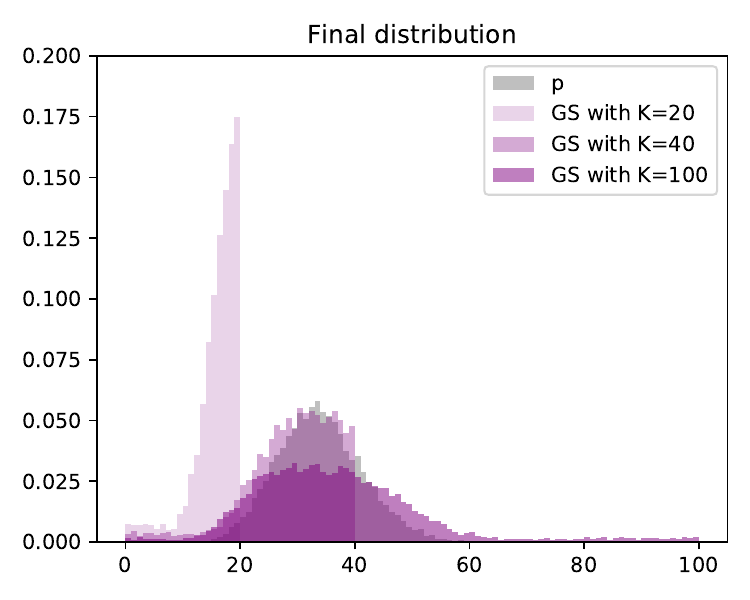}
  \caption{IGR-SB approximation to a Negative Binomial$(r=50,
  p=0.6)$.}\label{fig:negative_binomial}
 \end{figure}

Here again we see how the GS has difficulty approximating another distribution with a countably
infinite support. The GS with $K=40$ (middle-purple) doest not assign mass to the right tail where
as the GS with $K=100$ has difficulty taking out sufficient weight from the right tail of the
distribution.

\end{document}